\documentclass[letterpaper]{article} 
\usepackage[dvipsnames]{xcolor}
\usepackage{tabularx}
\usepackage[]{aaai24}  
\usepackage{times}  
\usepackage{booktabs}
\usepackage{helvet}  
\usepackage{courier}  
\usepackage[hyphens]{url}  
\usepackage{graphicx} 
\usepackage{natbib}  
\usepackage{caption} 
\usepackage{algpseudocode}
\usepackage{multirow}
\usepackage{array, bigdelim, makecell} 

\usepackage[ruled,vlined]{algorithm2e}
\usepackage{amsmath}
\title{Neural Video Compression with Temporal Layer-Adaptive Hierarchical B-frame Coding}
\author{
    Yeongwoong Kim\textsuperscript{\rm 1},
    Suyong Bahk\textsuperscript{\rm 1},
    Seungeon Kim\textsuperscript{\rm 2},
    Won Hee Lee\textsuperscript{\rm 2},
    Dokwan Oh\textsuperscript{\rm 2},
    Hui Yong Kim\textsuperscript{\rm 1}
}
\affiliations{
    \textsuperscript{\rm 1}School of Computing, Kyung Hee University, Republic of Korea\\
    \textsuperscript{\rm 2}Samsung Advanced Institute of Technology, Republic of Korea\\

}

\usepackage{bibentry}

\begin{document}

\maketitle

\begin{abstract}
Neural video compression (NVC) is a rapidly evolving video coding research area, with some models achieving superior coding efficiency compared to the latest video coding standard Versatile Video Coding (VVC). In conventional video coding standards, the hierarchical B-frame coding, which utilizes a bidirectional prediction structure for higher compression, had been well-studied and exploited. In NVC, however, limited research has investigated the hierarchical B scheme. In this paper, we propose an NVC model exploiting hierarchical B-frame coding with temporal layer-adaptive optimization. We first extend an existing unidirectional NVC model to a bidirectional model, which achieves -21.13\% BD-rate gain over the unidirectional baseline model. However, this model faces challenges when applied to sequences with complex or large motions, leading to performance degradation. To address this, we introduce temporal layer-adaptive optimization, incorporating methods such as temporal layer-adaptive quality scaling (TAQS) and temporal layer-adaptive latent scaling (TALS). The final model with the proposed methods achieves an impressive BD-rate gain of -39.86\% against the baseline. It also resolves the challenges in sequences with large or complex motions with up to -49.13\% more BD-rate gains than the simple bidirectional extension. This improvement is attributed to the allocation of more bits to lower temporal layers, thereby enhancing overall reconstruction quality with smaller bits. Since our method has little dependency on a specific NVC model architecture, it can serve as a general tool for extending unidirectional NVC models to the ones with hierarchical B-frame coding.

\end{abstract}

\section{1. Introduction}

Neural video compression (NVC) is a rapidly evolving field within video coding research, showcasing its potential to surpass the coding efficiency of established video coding standards such as High Efficiency Video Coding (HEVC)~\cite{sullivan2012overview} and Versatile Video Coding (VVC)~\cite{bross2021overview, xiang2022mimt, li2023neural}. However, while traditional video coding standards have extensively studied hierarchical B-frame coding, which employs bidirectional prediction structures to achieve higher compression, the exploration of this scheme in NVC has been limited~\cite{jvet2023algorithm}. In this paper, we propose an NVC model that leverages hierarchical B-frame coding and introduces temporal layer-adaptive optimization to enhance the compression efficiency of bidirectional NVC models. 

We first present a bidirectional extension of an existing unidirectional NVC model, deep contextual video compression (DCVC)~\cite{li2021deep}, which we term bidirectional DCVC (Bi-DCVC). This model includes bidirectional motion estimation, bidirectional motion coding, and bidirectional context generation and achieves -21.13\% BD-rate gain against DCVC. Despite achieving significant improvements over unidirectional baselines, Bi-DCVC shows significant performance degradation when applied to sequences containing complex or large motions. Based on the analysis results, we introduce a temporal layer-adaptive optimization strategy for bidirectional NVC models. More specifically, temporal layer-adaptive quality scaling (TAQS) and temporal layer-adaptive latent scaling (TALS) are proposed. The first one is a training strategy that employs distinct loss functions for different temporal layers, while the second is a method that scales latent features using temporal layer-wise scaling vectors. This allows for different operations for the temporal layers in a single model with small additional components.


\begin{figure}[!]
\centering
\includegraphics[width=.95\linewidth]{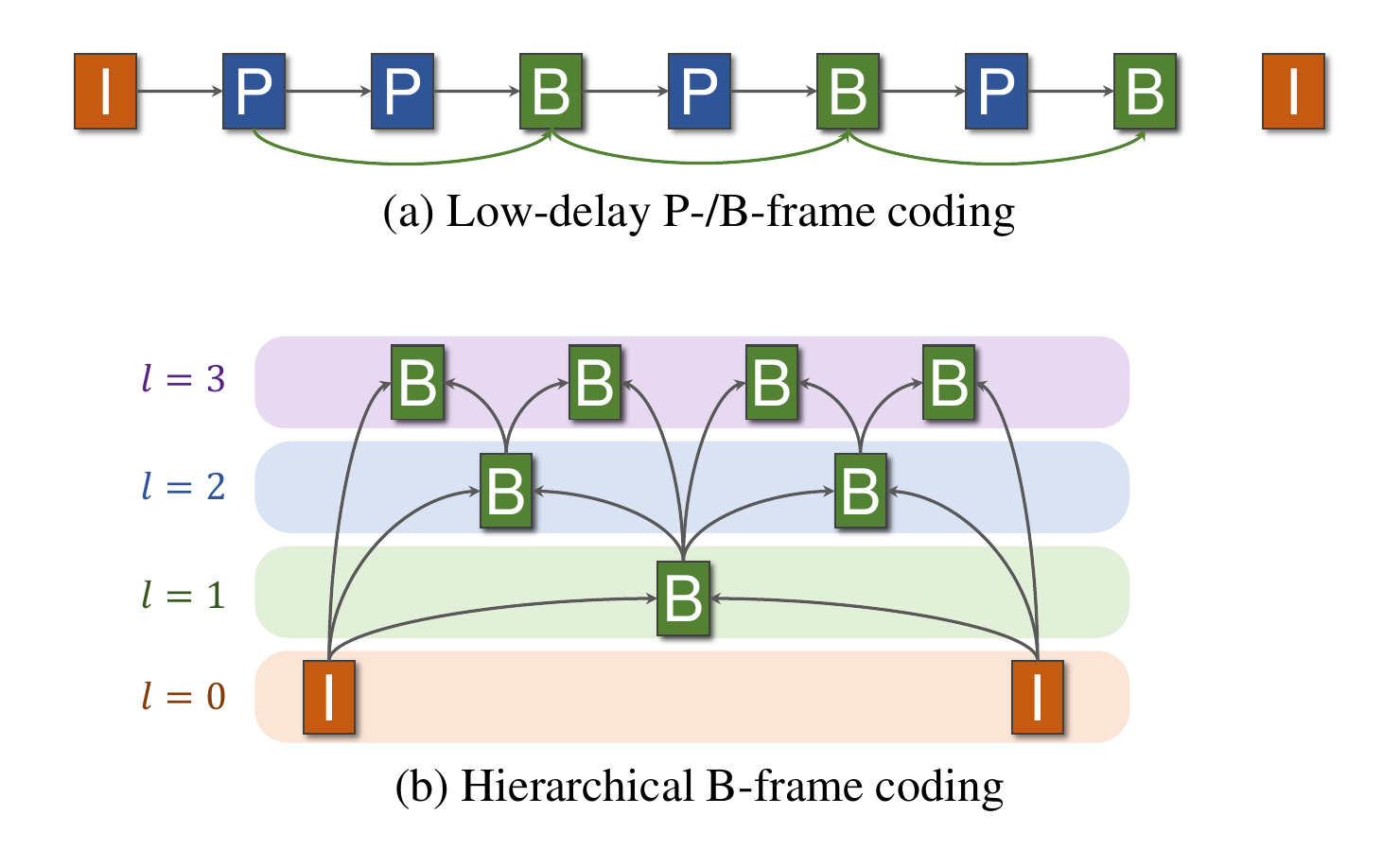} 
\caption{Illustration of the prediction structures in low-delay P/B and random access configurations.}
\label{fig1}
\end{figure}

We designate a model that integrates these two methods atop Bi-DCVC as hierarchical DCVC (Hi-DCVC). Hi-DCVC demonstrates improved coding efficiency compared to Bi-DCVC with a -39.86\% BD-rate gain against DCVC. In addition, for sequences with large or complex motions, Hi-DCVC shows up to -49.13\% more BD-rate reduction than Bi-DCVC. These improvements are attributed to the integration of temporal layer-adaptive optimization, which allocated more bits to lower temporal layers, resulting in improved reconstruction quality with fewer bits. Overall, Hi-DCVC's utilization of this optimization technique yielded impressive coding performance enhancements.

The main contributions of this paper include identifying challenges in extending unidirectional NVC models to bidirectional prediction structures and proposing novel strategies for temporal layer-adaptive optimization. The experimental results validate the effectiveness of our methods, showcasing significant improvements in coding efficiency and offering new avenues for enhancing bidirectional NVC models.

\begin{figure*}[!]
\centering
\includegraphics[width=.95\textwidth]{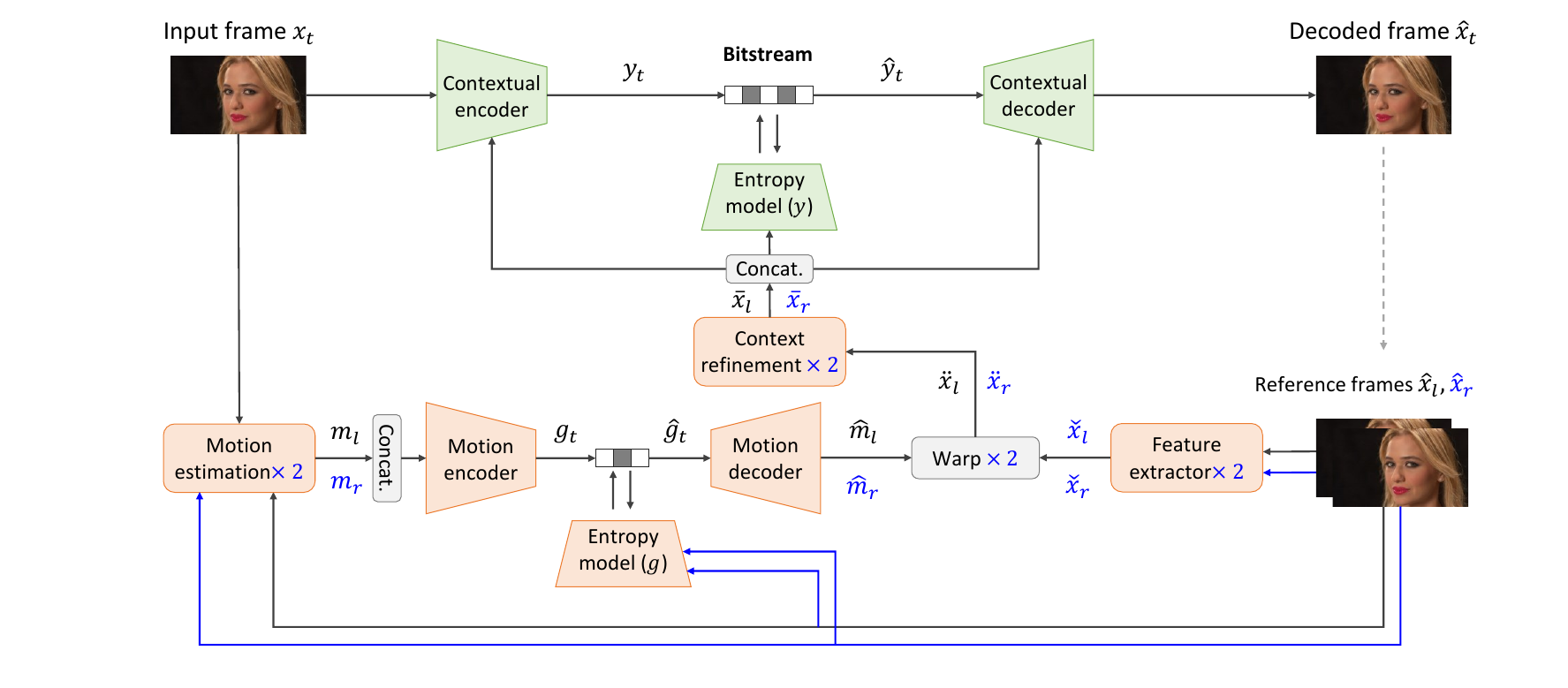} 
\caption{Overall architecture of the proposed Bi-DCVC with the differences from DCVC being highlighted in blue.}
\label{fig2}
\end{figure*}

\section{2. Related Works}

\subsection{2. 1. Hierarchical B-frame Coding}
There are various picture types, such as I-frame, P-frame, and B-frame, based on their prediction structure in traditional video coding. I-frame is a picture type that uses intra-coding without referencing other frames. P-frame and B-frame reference decoded frames to utilize temporal redundancy and can achieve higher coding efficiency than I-frame. P-frame uses only one reference frame, while B-frames can use two or more reference frames, as shown in Fig.~\ref{fig1}. 

While such inter-prediction structures yield high coding efficiency, they have coding delays contingent upon the referencing structures. Hence, video coding standards, such as HEVC~\cite{sullivan2012overview} and VVC~\cite{bross2021overview}, provide various configurations tailored to different applications~\cite{jvet2023algorithm}. For instance, the Low-Delay P/B (LDP/LDB) configuration aligns frames' output order and decoding order, allowing reference to only preceding frames, as shown in Fig.~\ref{fig1}-(a). These configurations are mainly used in real-time broadcasting and video conferencing applications. On the other hand, the Random Access (RA) configuration, incorporating hierarchical B-frame coding, enables referencing both preceding and subsequent frames based on the output order, attaining the highest coding efficiency. However, this configuration also entails the highest coding delays where it is more suitable for streaming services such as YouTube and Netflix.

Within the hierarchical B-frame coding scheme depicted in Figure~\ref{fig1}-(b), the number of temporal layers ($l=0,1,2,3$) is determined based on referencing structures. Although the illustration showcases a dyadic prediction structure with an intra-period of 8, it can be expanded to accommodate any intra-period that is a power of 2 with more temporal layers. The concept of temporal layers not only offers various functionalities, including scalable coding but also enhances coding efficiency. For example, frames in the lowest temporal layers are referenced most by other frames, making it advantageous to allocate more bits to the layer. In other words, by adjusting the bit allocation differently for different temporal layers, a substantial improvement in overall coding efficiency can be achieved.

\subsection{2.2. Neural Video Compression}
\subsubsection{Unidirectional NVC Models}
The success of neural image compression has paved the way for neural video compression, starting with the introduction of deep video compression (DVC) framework~\cite{lu2019dvc} which uses an unidirectional prediction structure. In the DVC framework, all the components of traditional video coding, such as motion estimation, motion compensation, and residual compression, are replaced with neural network modules, allowing for end-to-end training. Subsequent works improved the framework by introducing reference smoothing with scale-space flow~\cite{agustsson2020scale}, recurrent use of multiple reference frames~\cite{yang2021learning}, and more lightweight and enhanced network architectures~\cite{rippel2021elf}.

Furthermore, research has also delved into exploring new frameworks that harness feature domains with richer information than the image domain~\cite{hu2021fvc, li2021deep, li2022hybrid, li2023neural}, For instance, FVC~\cite{hu2021fvc} performs motion estimation and compensation from shallow features, utilizing offsets of deformable convolutions~\cite{dai2017deformable} as motion cues in the decoder, rather than bilinear warping based on optical flows. In addition, DCVC~\cite{li2021deep} proposed a contextual coding approach wherein shallow features extracted from reference frames are utilized to generate temporal with motion compensation. This context is then used as additional input to the encoder, decoder, and entropy model, introducing contextual coding that deviates from the residual coding scheme transmitting only the difference between predicted and current frames. Recently, NVC models using transformers~\cite{mentzer2022vct} or masked image modeling~\cite{xiang2022mimt} have also emerged. Surprisingly, DCVC-HEM~\cite{li2022hybrid} and DCVC-DC~\cite{li2023neural} building upon DCVC~\cite{li2021deep} have achieved higher compression performance compared to VVC~\cite{bross2021overview}, through additional research on temporal context and enhancements to the entropy model.

\subsubsection{Bidirectional NVC Models}
While the majority of models in neural video compression employ unidirectional prediction structures, some have explored the benefits of bidirectional prediction~\cite{yang2020learning, pourreza2021extending, yilmaz2021end, cetin2022flexible}, which is the main focus of this paper. For example, HLVC~\cite{yang2020learning} divides a group of pictures (GOP) into three different layers and  uses different coding methods and quality levels for the layers to exploit the advantages of bidirectional prediction. B-EPIC~\cite{pourreza2021extending} extends a unidirectional NVC model to a bidirectional one by a frame interpolation method, achieving high compression efficiency. LHBDC~\cite{yilmaz2021end} proposed novel bidirectional tools such as temporal bi-directional prediction of motion vectors and learned bi-directional motion compensation mask model for bidirectional neural video coding.

However, despite the advancements enabled by these bidirectional NVC models, certain limitations still persist. For instance, while HLVC utilizes different coding methods based on temporal layers, it is mainly focused on only two temporal layers, thus not providing a generic approach for temporal layer-adaptive optimization. Additionally, this approach requires storing multiple models, leading to memory overhead and loading latency. In addition, while B-EPIC presents a simple and effective method for extending unidirectional models to bidirectional ones, it lacks a study for temporal layers. As a result, it outperforms the unidirectional model with only around -22\% of BD-rate gains. LHBDC also neglects the temporal layer aspect. Motivated by these limitations, we conduct temporal layer analysis for bidirectional models and propose temporal layer-adaptive optimization methods in the following sections.

\begin{table*}[th!]
\centering
\resizebox{0.95\linewidth}{!}{
    \begin{tabular}{c c c c c c c c c c c}
    \toprule
    & & \multicolumn{8}{c}{\textbf{BD-rate (\%) gains ($\downarrow$)}} \\ 
    \cmidrule{3-10}
    
     & Intra-period & {Beauty} & {Bosphorus} & {HoneyBee} & {Jockey} & {ReadySteadyGo} & {ShakeNDry} & {YachiRide} & \textbf{Average} \\
    
    \midrule
    
    \multirow{2}{*}{\textbf{Bi-DCVC}} & 16 & \textcolor{Mahogany}{+50.57} & -44.99 & -36.98 & -2.51 & -3.17 & -18.23 & -13.23 & \textbf{-18.93} \\
    
    \cmidrule{2-10}

    & 32 & \textcolor{Mahogany}{+30.19} & -44.54 & -62.53 & \textcolor{Mahogany}{+6.79} & \textcolor{Mahogany}{+7.89} & -29.05 & -9.78 & \textbf{-21.13} \\
    
    \midrule
    
    \multirow{2}{*}{\textbf{Hi-DCVC (w/o TALS)}} & 16 & \textcolor{Mahogany}{+7.12} & -48.71 & -37.80 & -42.94 & -27.30 & -13.99 & -26.98 & \textbf{-32.47} \\
    \cmidrule{2-10}
    
    & 32 & -7.54 &		-54.99 &		-63.39& 		-40.81 &		-21.89 &		-24.74 &		-27.62 & \textbf{-37.72}	\\
    
    \midrule
    
    \multirow{2}{*}{\textbf{Hi-DCVC}} & 16 & \textcolor{Mahogany}{+1.58} & -50.55 & -38.03 & -43.96 & -28.52 & -13.94 & -28.09 & \textbf{-34.43} \\
    \cmidrule{2-10}
    
    & 32 & -12.73 &		-57.12 &		-63.45& 		-42.34 &		-23.35 &		-25.48 &		-28.92 & \textbf{-39.86}	\\
    \bottomrule
    \end{tabular}
    }
\caption{BD-rate gains of Bi-DCVC and Hi-DCVC against DCVC in terms of PSNR.}
\label{tab1}
\end{table*}

\section{3. Bi-DCVC: Bidirectional Extension of DCVC}
\subsection{3.1. Model Architecture}
Our work aims to improve the performance of unidirectional NVC models by leveraging the advantages of hierarchical B-frame coding. We first reproduce the existing unidirectional NVC model, deep contextual video compression (DCVC)~\cite{li2021deep}, and further extend it to a bidirectional model, Bi-DCVC. Subsequently, we introduce Hi-DCVC that overcomes the limitations of Bi-DCVC through temporal layer-adaptive optimization methods. In this section, we explain the Bi-DCVC framework.

Similar to the DCVC, our bidirectional model, Bi-DCVC, also consists of motion estimation, motion encoding and decoding, context generation, and contextual encoding and decoding processes. Fig.~\ref{fig2} illustrates the overall architecture of the proposed Bi-DCVC, where the differences from DCVC are highlighted in blue. Bi-DCVC uses two reference frames $\{\hat{x}_l, \hat{x}_r\}$ to generate bidirectional contexts of an input frame. In the motion estimation stage, SPyNet~\cite{ranjan2017optical} is employed to estimate two motion vectors $m_l$ and $m_r$, separately. These two motion vectors are concatenated across channels and compressed by a motion encoder. The motion encoder transforms these inputs into latent representation $g_t$, which is uniformly quantized and then entropy coded by the mean-scale hyperprior entropy model~\cite{minnen2018}. $\hat{g}_t$ represents the quantized latent feature of motion vectors, and the motion decoder reconstructs motion vectors $\hat{m}_l$ and $\hat{m}_r$ from the latent feature.

In the motion part, most components remain unchanged from DCVC. However, in some modules, the number of channels in the first or last layer is adjusted if needed. In addition, we also use a temporal prior encoder for extracting bidirectional temporal prior from $\{\hat{x}_l, \hat{x}_r\}$ in the motion entropy model to save the bits for bidirectional motion. The network architecture of the temporal prior encoder is the same as the temporal prior encoder in DCVC~\cite{li2021deep}, except for the inputs and the number of channels of the first and last layer. The details of the model architecture of Bi-DCVC can be found in the supplementary material.

Similarly, the context generation and contextual coding parts have no significant changes from DCVC, where the warping process, feature extractor, and context refinement modules are performed twice each. The outputs of the context refinement module, $\{\bar{x}_l, \bar{x}_r\}$, are concatenated across channels and provided as inputs to the contextual encoder and decoder, as well as the entropy model. The contextual encoder transforms the input frame into a latent feature $y_t$ conditioned on the bidirectional contexts, and the contextual decoder reconstructs the input frame $\hat{x}_t$ from the decoded latent feature $\hat{y}_t$. The contexts are also utilized in entropy encoding and decoding as additional inputs to the mean-scale hyperprior entropy model~\cite{minnen2018}.

\begin{figure}[h!]
\centering
\includegraphics[width=\linewidth]{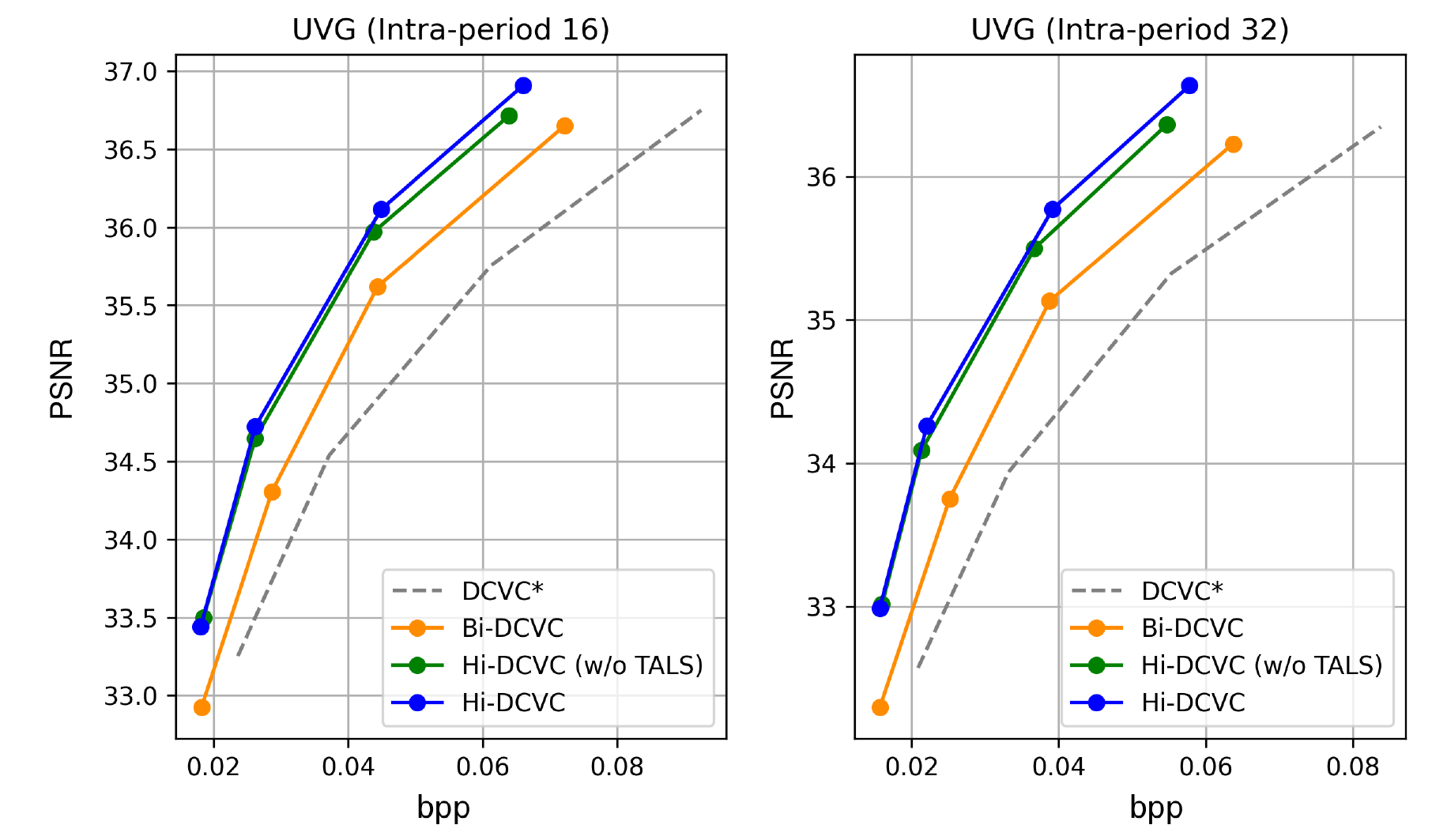} 
\caption{Rate-distortion curves of DCVC and the proposed models in this paper for UVG dataset. DCVC* is our reproduced model of DCVC.}
\label{fig3}
\end{figure}

\begin{figure*}[h!]
\centering
\includegraphics[width=.95\linewidth]{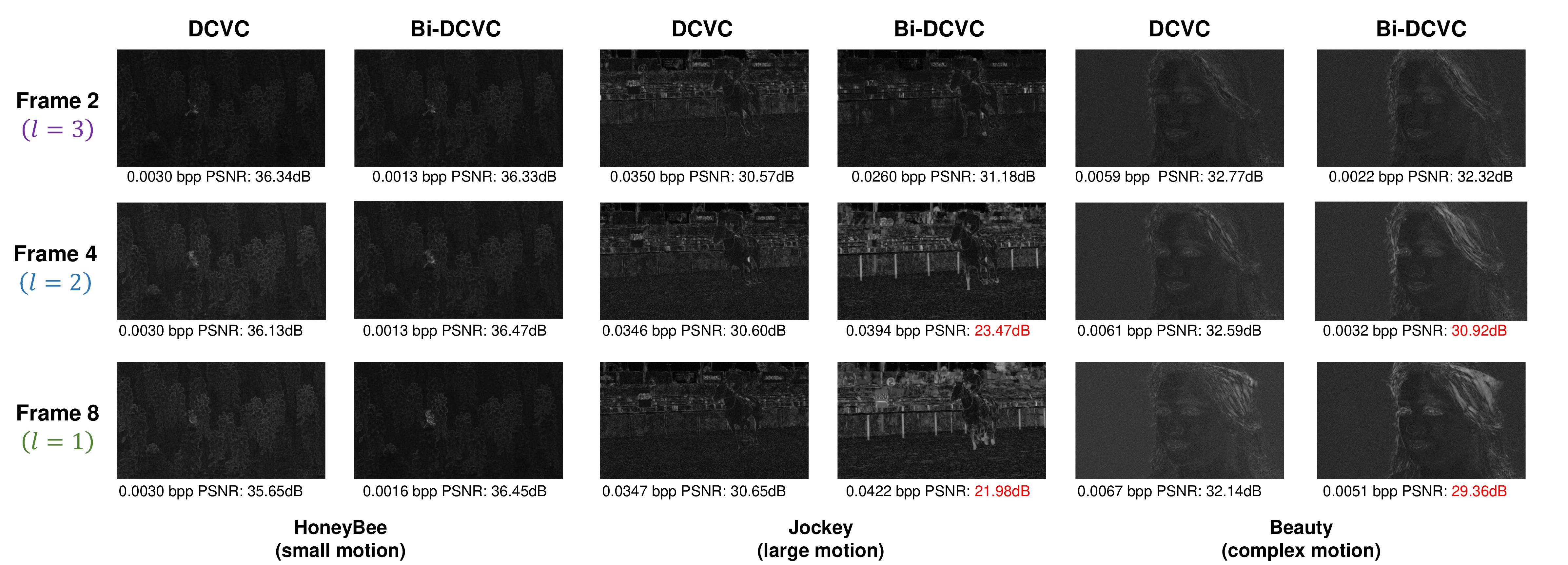} 
\caption{Visualization of the residual images from the contextual prediction of the ``HoneyBee'', ``Beauty'' and ``Jockey'' sequences by DCVC and Bi-DCVC.}
\label{fig4}
\end{figure*}

\begin{figure}[h!]
\centering
\includegraphics[width=.95\linewidth]{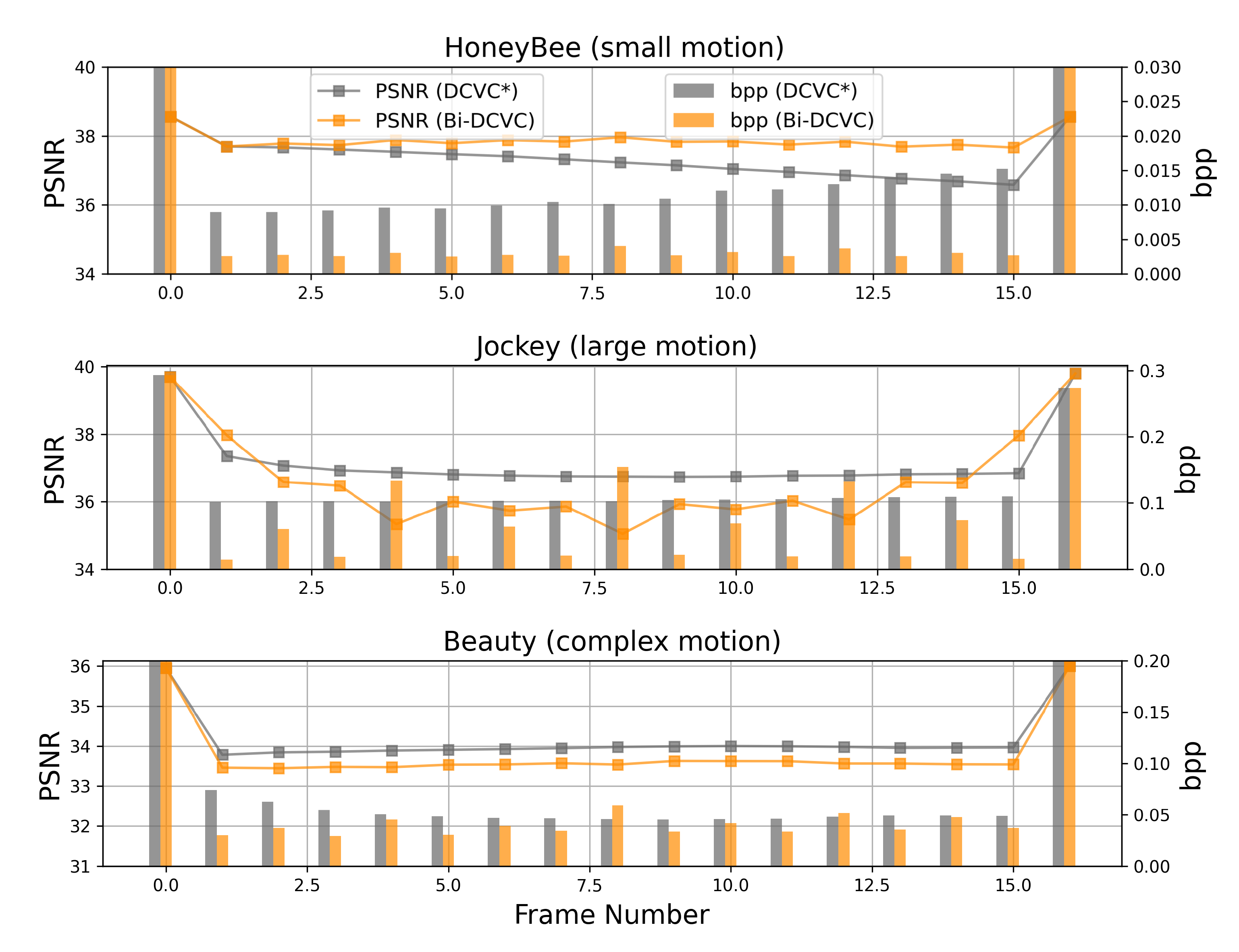} 
\caption{Comparisons of DCVC* and the proposed Bi-DCVC with frame-by-frame bit allocation and reconstruction quality for three sequences that have distinct motion types.}
\label{fig5}
\end{figure}

\subsection{3.2. Training Strategy}

Our model is trained with the Vimeo-90k dataset~\cite{xue2019video} which contains 91,701 video sequences. During training, we randomly crop each frame to $256 \times 256$ patch. The training sequences are composed of seven consecutive frames each, and we randomly sample five frames, as denoted by $\{x_0, ..., x_4\}$, to encompass various temporal distances. For I-frames, we compress $x_0$ and $x_4$ with an existing neural image codec~\cite{li2022hybrid}. Regarding the remaining frames, B-frame coding is performed. The prediction structure for the B-frames is determined by two predefined lists of triplets: [(0, 1, 4), (1, 3, 4), (1, 2, 3)] or [(0, 3, 4), (0, 1, 3), (1, 2, 3)]. Each triplet comprises three indices $(l, t, r)$, where $l$ and $r$ represent reference frame indices, and $t$ indicates the index of the frame to be encoded. For example, in the case of the first list, using $\hat{x}_0$ and $\hat{x}_4$ as reference frames, $x_1$ is coded in B-frame. Subsequently, with $\hat{x}_1$ and $\hat{x}_4$ as reference frames, $x_3$ is coded, followed by using $\hat{x}_1$ and $\hat{x}_3$ to code $x_2$. For each iteration, one of the two lists is randomly selected to determine the prediction structure. The difference between the two lists lies in the temporal distances between forward and backward references from the input frame. Employing these two lists for training aims to introduce diversity in prediction structures.

The loss function for each frame $x_t$ is defined as follows.
\begin{equation}
\label{eq1}
\begin{array}{cl}
    L_{t}=R(\hat{y}_t) + R(\hat{z}_t) + R(\hat{g}_t) + R(\hat{h}_t) & \\ \\ \quad \quad \quad \quad +  \lambda \cdot D(x_t, \hat{x}_t),&
\end{array}
\end{equation}
where $\hat{y}_t$ and $\hat{g}_t$ represent quantized latent feature of $x_t$ and bidirectional motion vectors $\{m_{l}$, $m_{r}\}$, respectively. Additionally, $\hat{z}_t$ and $\hat{h}_t$ denote the hyperprior latent of $\hat{y}_t$ and $\hat{g}_t$, respectively. We used mean squared error for the distortion $D(\cdot, \cdot)$ between input frame $x_t$ and reconstructed frame $\hat{x_t}$. The distortion term is multiplied by the Lagrangian multiplier $\lambda$ which controls the trade-off between rate and distortion. Note that we compress three B-frames per sequence, and the model parameters are updated once with the averaged losses for each batch of sequences. We trained four models with a list of $\lambda$ values, $\Lambda=[117, 227, 435, 845]$ to support multiple quality levels.  We use batch size 4 and Adam optimizer~\cite{kingma2014adam}, where the initial learning rate is $10^{-4}$ until the validation loss becomes a plateau. Finally, the model is fine-tuned with a learning rate of $10^{-5}$ for 50K steps.

\subsection{3.3. Experimental Results of Bi-DCVC}
\label{sec:performance}
We used DCVC as the baseline model to evaluate our model, and for a fair comparison, we reproduced the DCVC model without the autoregressive context model~\cite{minnen2018}. Detailed explanations of the reproduced DCVC can be found in the supplementary material. To evaluate the compression performance of DCVC and Bi-DCVC, we used 97 frames from seven sequences in the UVG dataset~\cite{uvg}, with intra-period 16 and 32 settings. For the Bi-DCVC model, we adopted a dyadic prediction structure similar to the one depicted in Fig.~\ref{fig1}-(b). For instance, in scenarios using an intra-period of 16, four temporal layers ($l$=1, 2, 3, 4) were used for B-frame coding. Similarly to this, we used five temporal layers for intra-period 32.

The rate-distortion curves of DCVC and Bi-DCVC are shown in Fig.~\ref{fig3}, and Bi-DCVC has superior performance to DCVC. Table~\ref{tab1} provides BD-rate gains for various video sequences and average gains, wherein Bi-DCVC achieves averaged -18.93\% and -21.13\% BD-rate gains against DCVC in terms of PSNR with intra-period 16 and 32, respectively. However, as can be seen in the table, relatively lower gains or even losses are observed for sequences containing large motions such as ``Jockey'' or ``ReadySteadyGo''. Furthermore, in the ``Beauty'' sequence with unpredictable hair-blowing motions, Bi-DCVC experiences significant performance degradation, up to a 50.57\% BD-rate increase.

\subsection{3.4. Temporal Layer Analysis}
To analyze the substantial performance degradation observed in some sequences, we first compare the prediction performance of Bi-DCVC with DCVC across different temporal layers. Secondly, we examine the frame-by-frame bit allocation and reconstruction quality for each temporal layer. We use the setting of intra-period 16 for this analysis.

\subsubsection{Prediction Performance}
In traditional motion-compensated residual coding schemes, the difference between the predictive frame and input frame can be visualized to represent prediction performance. However, in models utilizing contextual coding schemes~\cite{li2021deep, li2022hybrid, sheng2022temporal, li2023neural}, where temporal context is used as predictive feature maps instead of a predictive frame, the conventional visualization methods cannot be directly applied. Thus, we introduce a contextual prediction method where the estimated mean $\mu_{t}$ of the latent feature $\hat{y}_t$ is used as an input to the contextual decoder instead of $\hat{y}_t$ itself to generate a predictive frame. This approach enables the comparison of prediction performance between models based on contextual coding.

Fig.~\ref{fig4} shows the residual images from the contextual prediction of the ``HoneyBee'', ``Beauty'' and ``Jockey'' sequences. 
Note that the bpp in this figure accounts only for the information used to generate the contextual prediction (\emph{i.e.}, bit amounts of $\hat{g}_t$, $\hat{h}_t$, and $\hat{z}_t$). Likewise, the PSNR in this figure is calculated with the contextual predictive frame against the original image. For the ``HoneyBee'' sequence, higher prediction PSNR is achieved even with fewer bits than DCVC, demonstrating bidirectional prediction's effectiveness for small motions. However, in ``Jockey'' and ``Beauty'' sequences which have large or complex motions, Bi-DCVC exhibits significantly lower prediction performance compared to DCVC, especially for the lower temporal layers.


\subsubsection{Frame-by-Frame Performance}
Fig.~\ref{fig5} shows the frame-by-frame bit allocation and reconstruction quality. For the ``HoneyBee'' sequence with small motions, higher reconstruction quality with fewer bits than DCVC is observed regardless of temporal layers. However, for the ``Jockey'' sequence with large motions, the significantly lower reconstruction quality is observed in the first and the second temporal layer frames (\emph{i.e.}, the 8th, 4th, and 12th frames) despite more bits being allocated than DCVC. This phenomenon appears to stem from the larger distance between the reference frames and the input frame. Lastly, for the ``Beauty'' sequence with complex motion, a relatively lower reconstruction quality even with more bits is observed in the first temporal layer frame. In this case, the bidirectional model does not yield any gains, resulting in BD-rate loss.

\section{4. Hi-DCVC: Temporal Layer-Adaptive Optimization}


As analyzed in Section 3.4, Bi-DCVC exhibits notable performance degradation for large or complex motion sequences, particularly in the low temporal layers. To counteract this, a potential solution involves allocating additional bits to the lowest temporal layer, thereby improving reconstruction quality and mitigating prediction errors stemming from large temporal gaps to the reference frames. In this regard, we introduce a temporal layer-adaptive optimization strategy for bidirectional NVC models. More specifically, temporal layer-adaptive quality scaling (TAQS) and temporal layer-adaptive latent scaling (TALS) are proposed. The first one is a training strategy that employs distinct loss functions for different temporal layers, while the second is a method that scales latent features using temporal layer-wise scaling vectors. This allows for different operations for the temporal layers in a single model with small additional components. We designate a model that integrates these two methods atop Bi-DCVC as hierarchical DCVC (Hi-DCVC).

\subsubsection{Temporal Layer-Adaptive Quality Scaling}
We replace the existing loss function in Eq.~(\ref{eq1}) with Eq.~(\ref{eq2}) to enable the lambda values to be assigned differently according to the temporal layer indices ($l=1,2,3$).

\begin{equation}
\label{eq2}
\begin{array}{cl}
    L_{t}=R(\hat{y}_t) + R(\hat{z}_t) + R(\hat{g}_t) + R(\hat{h}_t) & \\ \\ \quad \quad \quad \quad +  {\Lambda}[B + 2 - l] \cdot D(x_t, \hat{x}_t),&
\end{array}
\end{equation}
where $B$ denotes the base quality level ($B\in\{1,2,3,4\}$). For example, if $B=1$ and $l=3$, the smallest lambda value $\Lambda[0]$ will be chosen. Note that the number of base quality levels corresponds to the number of trained models. In order to use this strategy for all the base quality levels, we expand the existing list of lambda values to $\Lambda=[50, 117, 227, 435, 845, 1625]$. In addition, the predefined lists of triplets have been modified as follows: [(0, 2, 6), (2, 4, 6), (2, 3, 4)] and [(0, 4, 6), (0, 2, 4), (2, 3, 4)], in order to ensure that prediction structures with large temporal distances are always included in the training process.

\subsubsection{Temporal Layer-Adaptive Latent Scaling}
While employing distinct loss functions for different temporal layers to enable the model to learn bit allocation based on these layers is effective, achieving this solely with the same model parameters could be difficult. Therefore, we propose the concept of temporal layer-adaptive latent scaling, where the latent feature $y_t$ is scaled by a scaling vector $q_{l}$, which is trained for each temporal layer. The scaling process follows Eq.~(\ref{eq3}), and the scaled and quantized latent feature $\hat{y}^{s}_{t}$ is rescaled with the same scaling vector $q_{l}$ before being used as input to the decoder network.

\begin{equation}
\label{eq3}
\begin{split}
&    y^{s}_{t}[h][w][c] = y_{t}[h][w][c] \times {1\over{q_{l}[c]}} \times {1\over{Q[h][w][c]}}, \\ \\
&    \hat{y}_{t}[h][w][c] = \hat{y}^{s}_{t}[h][w][c] \times q_{l}[c] \times Q[h][w][c], \quad l=1, 2, 3
\end{split}
\end{equation}
where $h, w$, and $c$ represent the horizontal, vertical, and channel indices of the latent feature, respectively. Besides $q_l$, we also employ the spatial-channel-wise quantization~\cite{li2022hybrid} to enhance adaptability to spatial positions of the input frame. The 3D quantization volume $Q$ is independent of temporal layers and obtained from the hyperprior decoder. We apply TALS to both latent features of bidirectional motions and of the input image. Note that in inference time, a deeper hierarchy may be used than the one used during training. In this case, we use applying the highest temporal layer used in the training stage for the deeper ones.

\begin{figure}[th!]
\centering
\includegraphics[width=\linewidth]{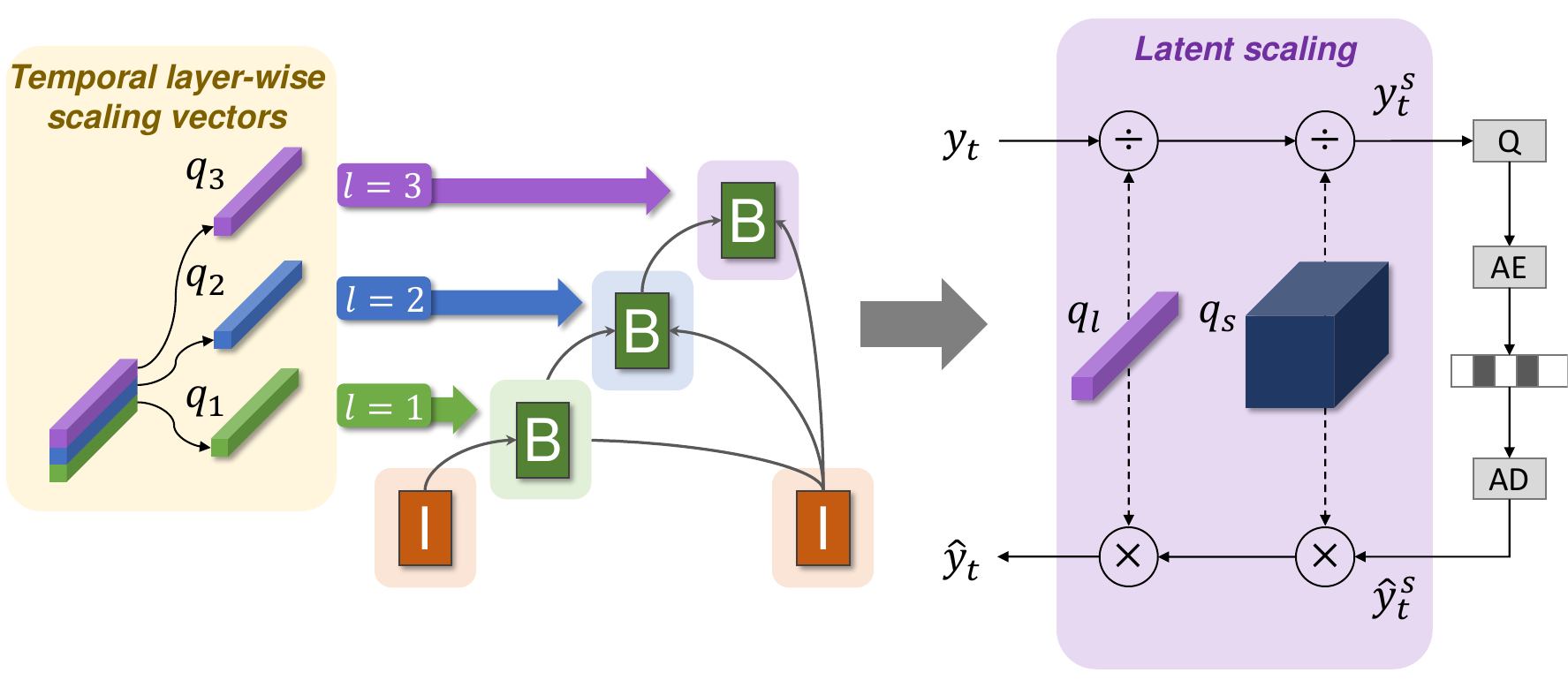} 
\caption{Illustration of the proposed temporal layer-adaptive latent scaling method combined with the spatial-channel-wise quantization~\cite{li2022hybrid}.}
\label{fig6}
\end{figure}

An illustration of the proposed TALS is shown in Fig.~\ref{fig6}. While implementing the proposed method, we employ a structure akin to multi-granularity quantization~\cite{li2022hybrid} for variable rate support. However, a notable distinction from previous research lies in the fact that while earlier studies utilized the same scaling vector for a given quality level, in ours, different scaling vectors are utilized based on the temporal layer even when referring to the same quality level.

\subsection{4. 2. Experimental Results of Hi-DCVC}
Using the same experimental settings as explained in the Bi-DCVC section, we evaluated the performance of Hi-DCVC. As shown in Fig.~\ref{fig3}, Hi-DCVC demonstrates improved coding efficiency compared to Bi-DCVC in rate-distortion sense. In Table~\ref{tab1}, Hi-DCVC shows -34.43\% and -39.86\% BD-rate gains against DCVC with intra-period 16 and 32, respectively. These experimental results indicate that by applying temporal layer-adaptive optimization to bidirectional NVC models, significantly improved performance can be achieved. Regarding the sequence-wise performance of Hi-DCVC, improvements are observed over Bi-DCVC for most sequences. Notably, for the sequences with large motion (``Jockey'' and ``ReadySteadyGO''), significant coding gains are shown with about 30-40\% BD-rate reductions, eliminating the BD-rate losses against DCVC. Similarly, for the ``Beauty'' sequence, which experienced significant BD-rate losses in Bi-DCVC, a majority of these losses are eliminated, even achieving a -12.73\% of BD-rate gain in the intra-period 32 setting.

\begin{figure}[t!]
\centering
\includegraphics[width=.95\linewidth]{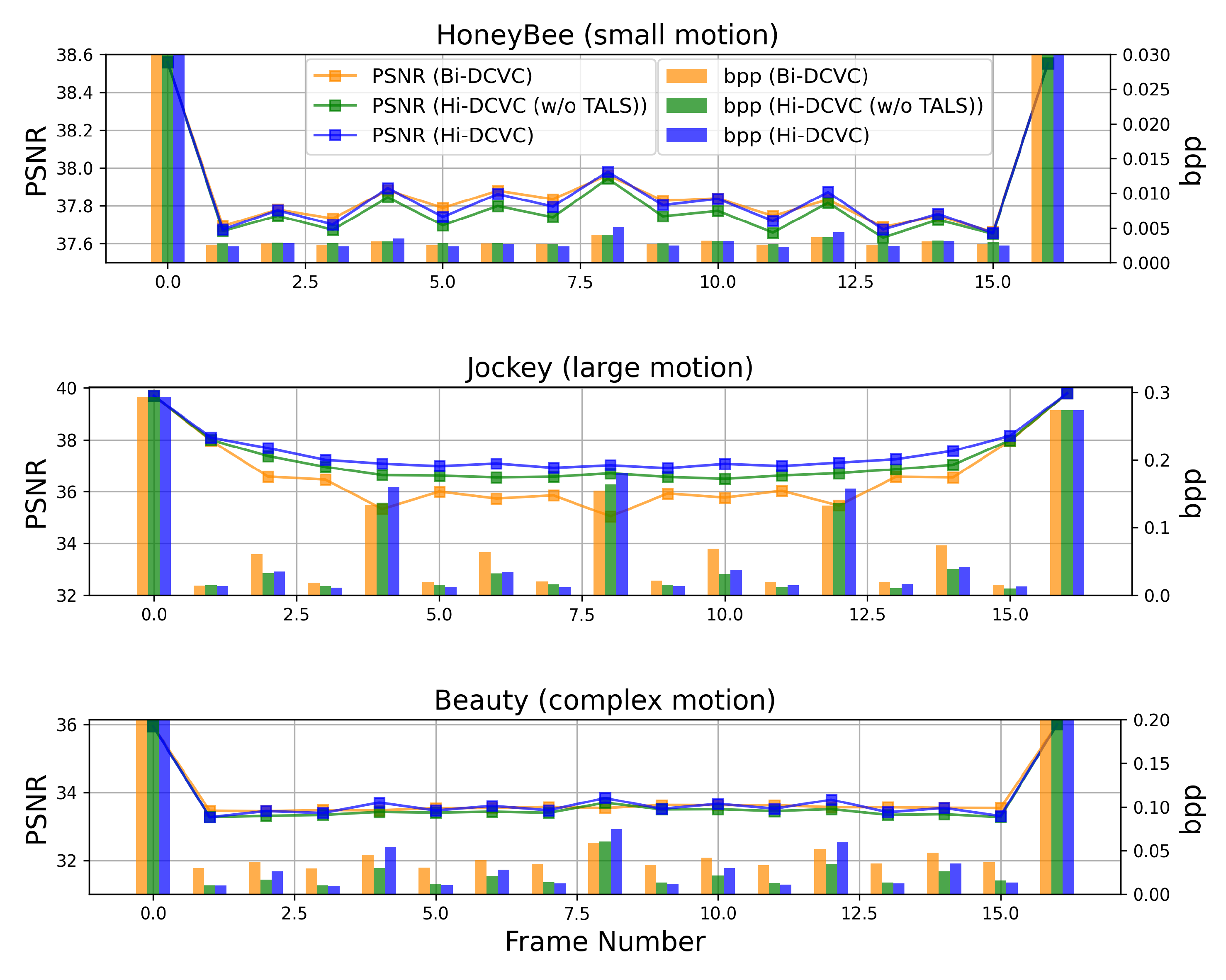} 
\caption{Comparisons of the proposed models, Hi-DCVC, Hi-DCVC, and Hi-DCVC (w/o TALS) with frame-by-frame bit allocation and reconstruction quality for three sequences that have distinct motion types.}
\label{fig7}
\end{figure}

Consistent with Section 3.4, we have visually depicted the frame-by-frame bit allocation and reconstruction quality for Hi-DCVC in Fig.~\ref{fig7}. As intended, with the utilization of temporal layer-adaptive optimization methods, we observe that across all sequences, more bits are allocated to the lower temporal layer frames, resulting in higher reconstruction quality. Furthermore, since the higher-quality frames in the lower temporal layers are referenced by the higher-layer frames, the lower-layer frames tend to consume fewer bits than Bi-DCVC with similar or higher reconstruction quality. Particularly, the overall PSNR improvement is remarkable in the ``Jockey'' sequence with large motions. Even in the case of the ``Beauty'' sequence with complex motions, Hi-DCVC improves the overall coding performance by minimizing bit consumption in the higher temporal layers.

\section{5. Conclusion}
In this study, we tackled the challenges of extending unidirectional NVC models to incorporate bidirectional prediction structures using hierarchical B-frame coding. The introduced Hi-DCVC model achieved significant coding gains over the baseline models with the proposed temporal layer-adaptive optimization methods. Since our methods have little dependency on a specific NVC model architecture, they can serve as a general tool for extending unidirectional NVC models to the ones with hierarchical B-frame coding. It is worth noting that while our study provides valuable insights, it was limited to training with three temporal layers. Future studies could explore the potential for extending our methods to accommodate deeper temporal layers.

\section{Acknowledgement}
This work is supported by Samsung Advanced Institute of Technology, Samsung Electronics Co., Ltd.

\clearpage

\section{Appendix}

\subsection{A1. Reproducing DCVC}

In order to apply our proposed method, we employed the unidirectional baseline DCVC~\cite{li2021deep} and reproduce this model. We re-wrote the model's implementation by referencing the official implementation available at \emph{``https://github.com/microsoft/DCVC''}, while the training code was developed by ourselves based on the DCVC paper. Although the components of the model remained nearly identical, we chose not to include the autoregressive context model~\cite{minnen2018} known for its substantial decoding time. For the sake of implementation convenience, we replaced the probability model for latent features from Laplacian distributions to Gaussian distributions. While DCVC originally employed the \textit{cheng2020-anchor} model from compressAI~\cite{begaint2020compressai} as an intra-coding model, this model also incorporated the autoregressive context model. Thus, we adopted the intra model proposed in ~\cite{li2022hybrid} without the context model.

\begin{figure}[h!]
\centering
\includegraphics[width=\linewidth]{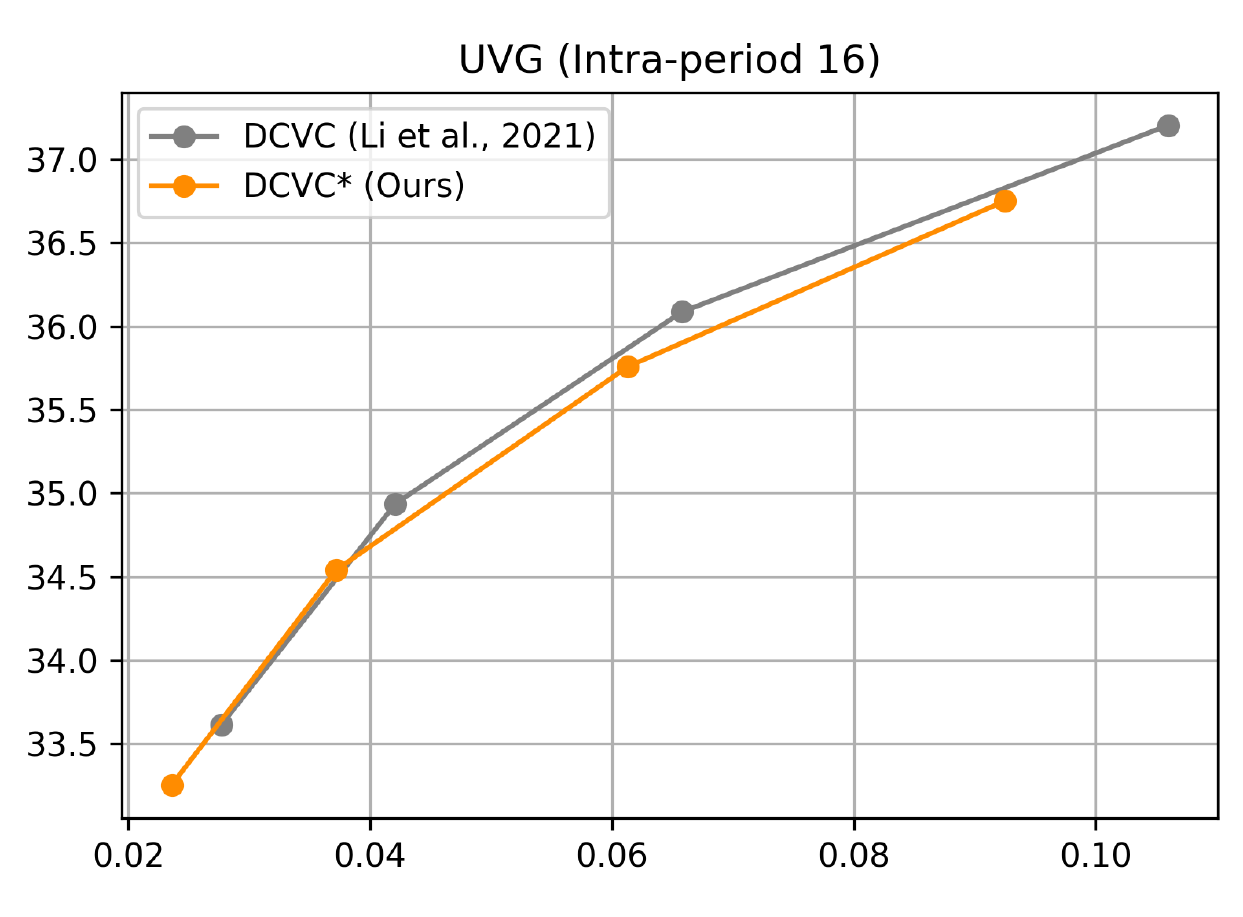} 
\caption{Rate-distortion curves of the original DCVC and our reproduced one.}
\label{sup_fig1}
\end{figure}

Fig.~\ref{sup_fig1} shows the performance comparison between the official implementation model and our reproduced one. Since we employed a distinct intra-coding model from the original DCVC, we used different lambda values [117, 227, 435, 845] from the lambda values [256, 512, 1024, 2048] used in the original DCVC. The omission of the autoregressive context model led to marginal performance degradation. Building upon the reproduced DCVC, we implemented our proposed models, Bi-DCVC and Hi-DCVC.

\begin{table*}[h!]
    \centering
    \resizebox{\linewidth}{!}{%
    \begin{tabular}{@{}cccccccc@{}}
    \toprule
    MV Encoder & MV Decoder & Contextual Encoder & Contextual Decoder & Feature Extractor & Context Refinement & \multicolumn{2}{c}{Temporal Prior Encoder} \\
    \midrule
    \textcolor{blue}{Conv 4, N*, k3, s2} & (Part 1) & \textcolor{blue}{Conv 2N+3, N, k5 s2} & (Part 1) & Conv N, N, k3, s1 & Resblock N, N, k3, s1 &  \textbf{[Image]} & \\
    GDN & Tconv N*, N*, k3, s2 & GDN & Subpixel, M, N, k3, s2 & Resblock N, N, k3, s1 & Conv N, N, k3, s1 & \textcolor{blue}{Conv 2N, N, k5, s2} & \\
    Conv N*, N*, k3, s2 & IGDN & Resblock N, N & IGDN & & & {GDN} & \\
    GDN & Tconv N*, N*, k3, s2 & GDN & Subpixel, N, N, k3, s2 & & & {Conv N, N, k5, s2} & \\
    Conv N*, N*, k3, s2 & IGDN & Resblock N, N & IGDN & & & {GDN} & \\
    GDN & Tconv N*, N*, k3, s2 & Conv N, N, k5, s2 & Resblock N, N & & & {Conv N, N, k5, s2} & \\
    Conv N*, N*, k3, s2 & IGDN & GDN & Subpixel, N, N, k3, s2 & & & {GDN} & \\
    & \textcolor{blue}{Tconv N*, 4, k3, s2} & Conv N, N, k5, s2 & IGDN & & &  {Conv N, N, k5, s2}& \\
    \cmidrule{2-2}
    \cmidrule{7-7}
    & (Part 2) & & Resblock N, N & & & \textcolor{blue}{\textbf{[Motion]}} & \\
    & \textcolor{blue}{Conv 10, N, k3, s1} & & Subpixel, N, N, k3, s2 & & &  \textcolor{blue}{Conv 6, N*, k5, s2} & \\
    \cmidrule{4-4}
    & LeackyReLU & & (Part 2) & & &  \textcolor{blue}{GDN} & \\
    & Conv N, N, k3, s1 & & \textcolor{blue}{Conv 3N, N, k3, s1} & & &  \textcolor{blue}{Conv N*, N*, k5, s2} & \\
    & LeackyReLU & & Resblock N, N & & &  \textcolor{blue}{GDN}& \\
    & Conv N, N, k3, s1 & & Resblock N, N & & &  \textcolor{blue}{Conv N*, N*, k5, s2}& \\
    & LeackyReLU & & Conv N, 3, k3, s1 & & &  \textcolor{blue}{GDN}& \\
    & Conv N, N, k3, s1 & & & & &  \textcolor{blue}{Conv N*, 2N*, k5, s2}& \\
    & LeackyReLU & & & & &  & \\
    & Conv N, N, k3, s1 & & & & &  & \\
    & LeackyReLU & & & & &  & \\
    & Conv N, N, k3, s1 & & & & &  & \\
    & LeackyReLU & & & & &  & \\
    & \textcolor{blue}{Conv N, 4, k3, s1} & & & & & & \\
    \bottomrule
    \end{tabular}%
    }
    \caption{Details of the network architectures for Bi-/Hi-DCVC. The differences from unidirectional DCVC are highlighted in blue. The layers are described as follows: (type, input channels, output channels, kernel size, stride). Note that N, M, and N* are set to 64, 96, and 128, respectively, as in DCVC. The terms "TConv" and "Subpixel" refer to transposed convolution and subpixel convolution layers~\cite{shi2016real}, respectively.}
    \label{sup_tab1}
\end{table*}

\begin{figure*}[h!]
\centering
\includegraphics[width=\linewidth]{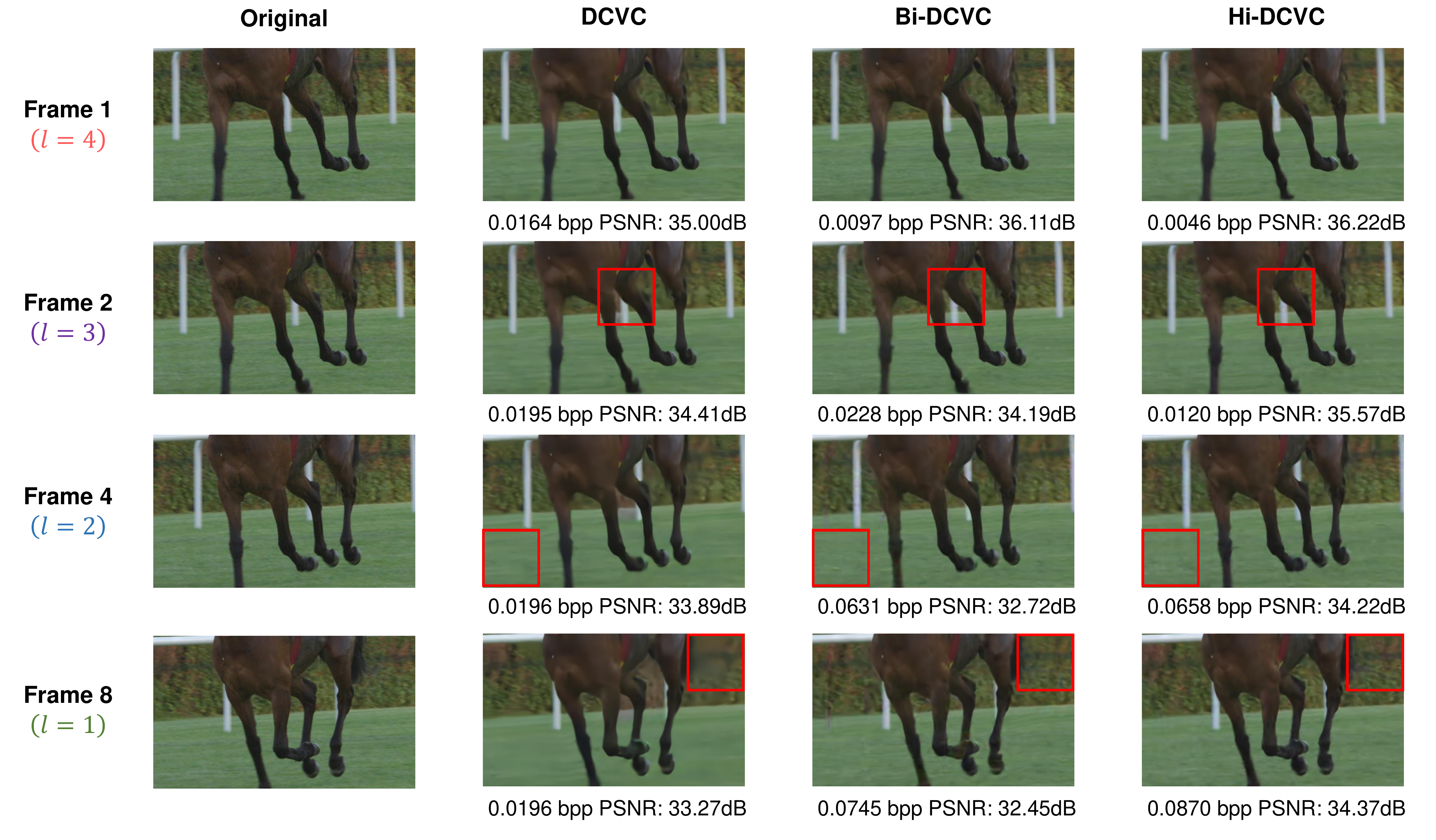} 
\caption{The visualization of the reconstructed frames by DCVC and the proposed models for qualitative comparison.}
\label{sup_fig2}
\end{figure*}

\subsection{A2. Detailed Network Architectures of Bi-DCVC and Hi-DCVC}
We aimed to enable bidirectional extension while making minimal modifications to the DCVC model, so we made adjustments to only a subset of layers from DCVC, as illustrated in Table~\ref{sup_tab1}. The differences from DCVC are highlighted in blue. Additionally, to extract a temporal prior for motion from bidirectional reference frames, which was absent in DCVC, we designed a temporal prior encoder similar to the one used in DCVC and incorporated it.

\subsection{A3. Qualitative Comparisons}
Fig. \ref{sup_fig2} provides a qualitative comparison between DCVC and our models. In the case of DCVC, it consumes similar bits for all frames, resulting in progressively lower reconstruction quality for later frames due to error propagation. In contrast, both Bi-DCVC and Hi-DCVC allocate bits differently based on the temporal layers ($l=1,2,3,4$). For Bi-DCVC, even with relatively low bits allocated to the first and second frames referencing the high-quality I-frame (frame 0), it exhibits comparatively high reconstruction quality. However, in lower temporal layers, despite using more bits than DCVC, the reconstruction quality is significantly lower. In Hi-DCVC, a slightly higher allocation of bits is assigned to the lower temporal layers than in Bi-DCVC, improving reconstruction quality for all frames. Notably, for the first and second frames, Hi-DCVC achieves much higher reconstruction quality with significantly fewer bits than DCVC. These outcomes highlight improved preservation of details in distant foliage and grass as well as a reduction in the green-colored artifacts present on the horse's hooves.

\bigskip

\bibliography{aaai24}

\end{document}